\documentclass[runningheads]{llncs}
\usepackage{graphicx}
\usepackage{array} 

\begin{document}

\title{Multi-Institutional Deep Learning Modeling Without Sharing Patient Data: A Feasibility Study on Brain Tumor Segmentation}

\titlerunning{Multi-Institutional Deep Learning Modeling Without Sharing Patient Data}

\author{Micah J Sheller\inst{1,3} \and
G Anthony Reina\inst{1} \and
Brandon Edwards\inst{1} \and
Jason Martin\inst{1} \and
Spyridon Bakas\inst{2,3}\orcidID{0000-0001-8734-6482}}

\authorrunning{M. J. Sheller et al.}

\institute{Intel Corporation, Santa Clara, CA 95052, USA \and
Center for Biomedical Image Computing and Analytics (CBICA), University of Pennsylvania, Philadelphia, PA 19104, USA \and
Corresponding authors\\
\email{\{micah.j.sheller@intel.com, sbakas@upenn.edu\}}}

\maketitle              

\begin{abstract}
Deep learning models for semantic segmentation of images require large amounts of data. In the medical imaging domain, acquiring sufficient data is a significant challenge. Labeling medical image data requires expert knowledge. Collaboration between institutions could address this challenge, but sharing medical data to a centralized location faces various legal, privacy, technical, and data-ownership challenges, especially among international institutions. In this study, we introduce the first use of federated learning for multi-institutional collaboration, enabling deep learning modeling without sharing patient data. Our quantitative results demonstrate that the performance of federated semantic segmentation models (Dice=0.852) on multimodal brain scans is similar to that of models trained by sharing data (Dice=0.862). We compare federated learning with two alternative collaborative learning methods and find that they fail to match the performance of federated learning.

\keywords{Machine Learning \and Deep Learning \and Glioma \and Segmentation \and Federated \and Incremental \and BraTS.}
\end{abstract}

\section{Introduction}
Gliomas describe tumors of the central nervous system with vastly heterogeneous radiographic, histologic, and molecular landscape. There is mounting evidence that tumor subregions apparent at the radiographic level reflect various histologically distinct tumor subregions with different biological properties. Accurate segmentation of these subregions is considered the basis for extracting quantitative imaging features corresponding to specific anatomical regions, which when integrated using advanced computational approaches, can enable assessment of this radiographic heterogeneity, evaluation of disease properties, and correlation with treatment response, patient prognosis, and molecular characteristics \cite{thePhiIndex,jayashreeIDH,ericksonMGMT,hamedRecurrence,lukeSurvival}.

The brain tumor segmentation (BraTS) challenge describes a successful effort to create a publicly available multi-institutional dataset for benchmarking and quantitatively evaluating the performance of computer-aided segmentation algorithms \cite{bratsTmi,NatureTciaGlioma,segmentationsGbm,segmentationsLgg}. However, such centralization of data, notwithstanding multi-institutional collaborations, is challenging because a) data availability is more limited when compared with real-world/photographic imagery, and b) sharing data to a centralized location may be cumbersome, especially in international configurations, due to various legal, privacy, technical, and data ownership challenges \cite{digitizationInHealthcare,privacyOnMedicalData}.

Considering the difficulty of creating public centralized medical imaging datasets, this paper introduces the first use of \emph{federated learning} (FL) \cite{arxivDlDecentralized} for medical imaging. Specifically, we apply FL on the BraTS data to build an effective segmentation model that learns the variation across multiple institutions, without sharing any patient data, by iteratively aggregating locally-trained models at a centralized server. Although we applied FL to supervised semantic segmentation using a deep convolutional neural network (CNN) architecture, namely U-Net, FL works with any supervised machine learning (ML) architecture. Furthermore, we compare FL against two alternative collaborating learning techniques: \emph{institutional incremental learning} (IIL), where each institution trains the shared model in turn, and \emph{cyclic institutional incremental learning} (CIIL), which is IIL done in rounds with prescribed numbers of epochs \cite{jayashreeDistributed}. We find that IIL performs poorly compared to FL and CIIL, while CIIL is less stable and harder to validate than FL, resulting in an inferior model.

Prior demonstrations of FL have focused on either toy problems or end-user tasks \cite{arxivDlDecentralized,federatedClient,federatedBackdoor,recurrentLanguage}. Our work is the first demonstration of FL in the medical domain, for institution-level tasks, specifically applied on clinically-acquired data.

\section{Materials and Methods}

\subsection{Federated Learning (FL) Overview}
\begin{figure}
\begin{center}
\includegraphics[width=0.49\textwidth]{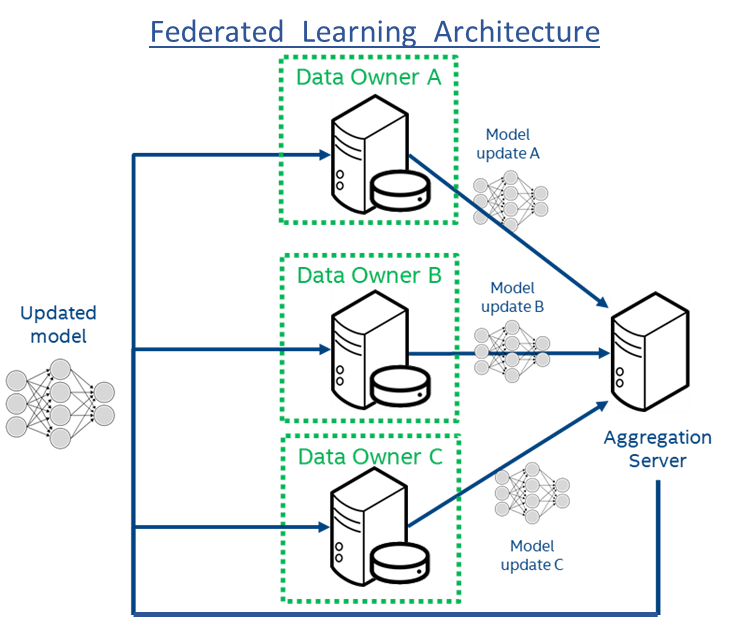}
\setlength{\abovecaptionskip}{-1pt}
\setlength{\belowcaptionskip}{-18pt}
\caption{System Architecture of Federated Learning.}
\label{fig1}
\end{center}
\end{figure}

In traditional ML solutions, all collaborating data owners (i.e., institutions) upload their data to a central server for training. Contrarily, in FL, the owners do not share their data, but they train the shared model locally instead, and only send model updates to the central server. The server then aggregates the updates and sends the new shared parameters to the data owners for further training (as often as desired), or application. Specifically, the aggregation is performed as a weighted average of institutional updates, with the weighting at a particular institution given as the fraction of total data instances that reside at that institution. Each iteration of this process: local training, update aggregation, and distribution of new parameters, is called a \emph{federated round} (Fig.\ref{fig1}).

\subsubsection{Hyper-Parameters and Convergence.}
Along with the typical hyper-parameters of deep learning architectures (e.g., batch size, optimizer, learning rate), FL also includes: a) epochs per round (EpR), b) number of participants in each round, and c) model update compression/pruning methods \cite{arxivDlDecentralized}. EpR influences convergence, as learning rate and batch size do in traditional training: i.e., more EpR can speed up convergence, but there are diminishing returns, especially when the institutions' datasets are not independent and identically distributed (IID) \cite{federatedNonIid}, and more EpR requires more compute per institution. We study this issue by experimenting with different sizes of simulated federations, from 4 up to 32 institutions, and training for various numbers of EpR.

Hyper-parameters of number of participants per round and update compression/pruning methods are not assessed in this study, as they are generally used to mitigate participant limitations that are less relevant in the medical domain (e.g., limited networks).
\subsubsection{Differential Privacy}
Though FL ensures raw data is never shared between collaborators, additional measures may be desired to prevent certain information from being obtained through model updates. For example, noise can be added before sending an update, to obscure the presence of any collection of samples in the institution's dataset, and an accounting can be made as to the likelihood that such a determination can be made from the resulting model. The model is then said to have a degree of `differential privacy'.

Differentially private training has been studied for ML use cases other than semantic segmentation \cite{privacyPreserving,differentialPrivacy}. However, applying noise to updates generally slows training, and there is a point at which training must stop to prevent increasing the likelihood of information leakage beyond an acceptable level. It is possible that the model has not reached the desired utility at that point. Differentially private ML models in the medical domain could be very desirable given the privacy issues surrounding medical data. However, we leave the study of differentially private training for segmentation models to future work.
\subsection{Institutional Incremental Learning (IIL)}
IIL is a simple collaborative learning approach, where institutions train a shared model in succession. Each institution trains the model only once, and may train the model however it chooses. Compared to FL, IIL requires less bandwidth as each institution needs to transmit the model once and receive it twice (once to train and once to receive the final version). The major disadvantages of IIL are a) the drop in performance as the number of institutions increases, and b) the problem of catastrophic forgetting \cite{catastrophicForgetting,overcomingForgetting}, where previously learned patterns are forgotten when new training data replace the previous data.

\subsection{Cyclic Institutional Incremental Learning (CIIL)}
CIIL changes IIL by repeating the IIL process, i.e., cycling through the institutions, and by fixing the number of epochs at each institution to reduce forgetting. In a CIIL cycle, each institution trains the model in series for a specific number of epochs before passing the updated model to the next participant. In contrast, during a federated round, each institution trains the model in parallel for a specific number of epochs, after which the institutions' model updates are aggregated to form the updated model. CIIL and FL share most of their software and infrastructure requirements, except for the FL aggregator. Although existing literature \cite{jayashreeDistributed} reports parallel collaborative training (e.g., FL) as more logistically complex, the results in the present study show that for CIIL to achieve comparable results to FL, CIIL requires additional validation overhead that makes it more complex and less efficient than FL.

\subsection{U-Net topology}
\begin{figure}
\begin{center}
\includegraphics[width=\textwidth]{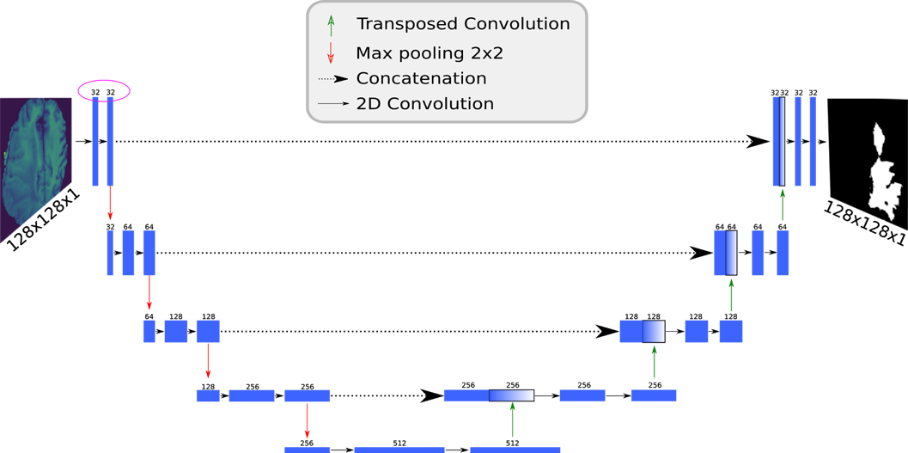}
\setlength{\abovecaptionskip}{-1pt}
\setlength{\belowcaptionskip}{-18pt}
\caption{U-Net network diagram. Numbers above each layer indicate the number of channels in that layer. Note that the channel count (purple circle) differs from the original design \cite{uNet} by a factor of 2.}
\label{fig2}
\end{center}
\end{figure}

For our analysis, we implemented\footnote{{https://github.com/NervanaSystems/topologies/tree/master/distributed\_unet}} a U-Net topology of a deep CNN \cite{uNet} (Fig.\ref{fig2}). The model takes as input a single channel image and outputs an equivalently-sized, binary mask in which each pixel is assigned a class label. The network mimics the architecture of an autoencoder, with a contracting path that captures context (via max pooling) and an expanding path that enables localization (via upsampling). Unlike the standard autoencoder, each feature map in the expanding path is concatenated with a corresponding feature map from the contracting path, augmenting downstream feature maps with spatial information acquired using smaller receptive fields. Intuitively, this allows the network to consider features at various spatial scales. Since its introduction in 2015, U-Net has quickly become one of the standard deep learning topologies for image segmentation and has been instrumental in creating prediction models for segmenting nerves in ultrasound images, lungs in CT scans, and even interference in radio telescopes. All of our collaborative learning experiments in this study used this model with a dropout parameter of 0.2 and upsampling set to true.

\subsection{BraTS Dataset}
To quantitatively evaluate FL in a medical imaging context, we used the BraTS 2018 training dataset \cite{bratsTmi,NatureTciaGlioma,segmentationsGbm,segmentationsLgg}, which contains multi-institutional multi-modal magnetic resonance imaging (MRI) brain scans from patients diagnosed with gliomas. The radiographically abnormal regions of each brain scan have been manually annotated using 3 distinct labels corresponding to i) peritumoral edematous/invaded tissue, ii) non-enhancing/solid and necrotic/cystic tumor core, and iii) enhancing tumor.

Since the objective of this study was to assess the performance of FL on a clinically-relevant task and not to develop a new segmentation method, we have only focused on the whole tumor volume, defined as the union of all three labels, only for the patients diagnosed with a high-grade glioma and we only considered the FLAIR modality for the input channel to the model.

\section{Experimental Results}
Our experiments compare traditional ML using data-sharing with collaborative configurations of FL, IIL and CIIL. For our collaboration experiments, we distribute the data among institutions in two different ways: 1) the actual BraTS distribution, i.e. the real-world data distribution, and 2) simulated distributions of 4 to 32 institutions, in steps of powers of two. The simulated distributions were created by randomly and proportionally splitting the subjects among the collaborating institutions, ensuring that each patient's data is assigned to only one institution. Table~\ref{table1} shows the average number of subjects per distribution for the simulated distributions, as well as the actual distribution of subjects across institutions. Note that the actual distribution is quite imbalanced, with a single institution contributing nearly half the data.

\begin{table}
\begin{center}
\caption{Distribution of data for all experimental configurations, simulated and real.}
\label{table1}
\begin{tabular}{|l|c|c|}
\hline
\textbf{Type of Distribution} & \textbf{Institutions} & \textbf{Average Subjects Per Institution}\\
\hline
Data-sharing & 1 & 178\\
Simulated & 4 & 44.5\\
Simulated & 8 & 22.2\\
Simulated & 16 & 11.1\\
Simulated & 32 & 5.6\\
Real (BraTS Distribution) & 10 & 70,27,17,12,11,9,6,6,4,3\\
\hline
\end{tabular}
\end{center}
\end{table}

For the `data-sharing' and `simulated' distributions, we randomly chose 32 subjects to hold out for validation on unseen data, prior to distributing the data among institutions. For the real BraTS distribution, we increased the unseen set to 45 subjects. This is because for institutions with only 4 and 5 subjects, contributing just 1 patient represents 20-25\% of their data, so we increased the unseen set proportion per institution to better balance their contributions. This does slightly penalize the results of the real distribution experiments compared to the data-sharing and simulated experiments.

Tables~\ref{table2} and \ref{table3} compare the data-sharing and three collaborative methods for the real and simulated distributions. For the data-sharing experiments, we show the best result from multiple model initializations. This matches normal practice for centralized training. Testing multiple model initializations may not be considered reasonable for collaborative methods, so we show the mean and standard deviation across multiple runs. For CIIL, we show results for all cycles over multiple runs, since CIIL does not provide a validation method for choosing the best cycle from a series of cycles. We discuss this further in section \ref{sec:bratsDistroResults}.

\begin{table}
\begin{center}
\caption{Data-sharing, FL, IIL and CIIL experiment results for the real BraTS data distribution.}
\label{table2}
\begin{tabular}{|c|c|c|}
\hline
\textbf{Method} & \textbf{Validation $DC$} & \textbf{Percent of Data-Sharing $DC$}\\
\hline
Data-sharing & 0.862 & 100\%\\
FL & 0.852$\pm$0.002 & 98.7\%\\
CIIL & 0.82$\pm$0.04 & 95\%\\
IIL & 0.803$\pm$0.042 & 93\%\\
\hline
\end{tabular}
\end{center}
\end{table}

\begin{table}
\begin{center}
\caption{Comparing FL, CIIL and IIL for collaborations of 4-32 institutions. Val.$DC$:Validation $DC$. D.S.$DC$:Percent of Data-Sharing $DC$.}
\label{table3}
\begin{tabular}{|c|c|c|c|c|c|c|}
\hline
\textbf{Institutions} & \multicolumn{2}{c|}{\textbf{FL}} & \multicolumn{2}{c|}{\textbf{CIIL}} & \multicolumn{2}{c|}{\textbf{IIL}}\\
 & Val.$DC$ & D.S.$DC$ & Val.$DC$ & D.S.$DC$ & Val.$DC$ & D.S.$DC$\\
\hline
4 & 0.862$\pm$0.003 & 99.9\% & 0.843$\pm$0.011 & 97.7\% & 0.841$\pm$0.004 & 97.4\%\\
8 & 0.865$\pm$0.002 & 100.2\% & 0.839$\pm$0.016 & 97.3\% & 0.823$\pm$0.014 & 95.4\%\\
16 & 0.863$\pm$0.002 & 99.9\% & 0.82$\pm$0.032 & 95\% & 0.82$\pm$0.018 & 95\%\\
32 & 0.857$\pm$0.001 & 99.3\% & 0.809$\pm$0.023 & 93.7\% & 0.701$\pm$0.058 & 81.20\%\\
\hline
\end{tabular}
\end{center}
\end{table}

\subsection{Benchmarking Metric}
The quantitative performance evaluation metric for the BraTS challenge has always been the Dice Coefficient ($DC$), a similarity measure in the range [0,1] that reflects a ratio of the intersection over the union of the predictions and ground truth, defined as:
\begin{equation}
DC=\frac{2 |P\cap T|}{|P|+|T|}
\end{equation}
where P and T are the prediction and ground truth masks, respectively.

The inter-rater agreement for expert neuro-radiologists measured by the $DC$ was reported in the original BraTS benchmark paper \cite{bratsTmi} equal to 0.85$\pm$0.08 (mean$\pm$std for the whole tumor segmentation). Furthermore, state-of-the-art models for this dataset have $DC$ of greater than or equal to 0.85 \cite{yongBrats}.

We used the Adam optimizer (learning rate 0.0005) to minimize the negative log of $DC$. To further increase numerical stability, we added a Laplace smoothing of 1 and we algebraically rearranged the final loss function to replace division with log subtraction:
\begin{equation}
loss=\log{(|P|+|T|+1)}-\log{(2|P\cap T|+1)}
\end{equation}

\subsection{Baseline U-Net Results}
\begin{figure}
\begin{center}
\includegraphics[width=0.7\textwidth]{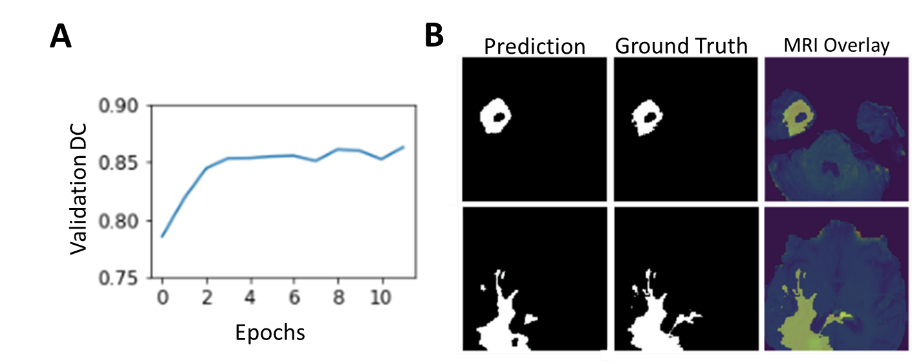}
\setlength{\abovecaptionskip}{-2pt}
\setlength{\belowcaptionskip}{-18pt}
\caption{(A) Validation $DC$ scores over training epochs. The model peaked at 12 epochs and achieves a validation $DC$ score of greater than 0.86. (B) Model performance on two images from the test set MRI. The model predicted mask closely matches the ground truth labels. An overlay of the ground truth with the original MRI slice.}
\label{fig3}
\end{center}
\end{figure}

The model trained to state-of-the-art accuracy within 3 epochs and reached a peak validation $DC$ of 0.862 (Fig.\ref{fig3}A) (15\% holdout data). A qualitative assessment of two MRI slices from the validation dataset shows a good agreement between the model predictions and the manually annotated boundaries (Fig.\ref{fig3}B).

\subsection{BraTS Distribution Results}
\label{sec:bratsDistroResults}
\begin{figure}
\begin{center}
\includegraphics[width=\textwidth]{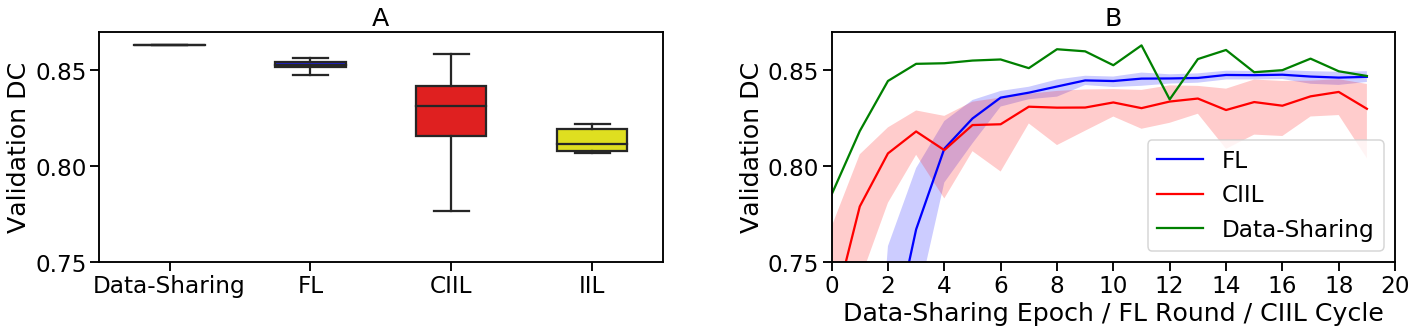}
\setlength{\abovecaptionskip}{-3pt}
\setlength{\belowcaptionskip}{-25pt}
\caption{Comparing centralized learning, FL, IIL and CIIL for the actual BraTS data distribution. The x-axis in (B) shows passes over the full dataset (epochs). Epochs are not equivalent in wall-clock time. The shading in (B) is min/max.}
\label{fig4}
\end{center}
\end{figure}

Figure \ref{fig4} shows comparative results across data-sharing, FL, IIL and CIIL for our U-Net implementation over the BraTS 2018 training dataset. In the collaborative experiments, the patient data was divided among the institutions exactly as it was collected by the BraTS data contributors. Figure~\ref{fig4}A shows how the scores vary across multiple runs, while Fig.\ref{fig4}B shows the validation score after each pass over the full training data. For the FL and CIIL experiments, the number of EpR was one. Note that in FL, each institution trains in parallel and the updates are averaged, such that the effective learning rate is less than that of the other methods. Furthermore, because FL trains in parallel, FL rounds and CIIL cycles are not wall-clock-equivalent.

\begin{figure}
\begin{center}
\includegraphics[width=\textwidth]{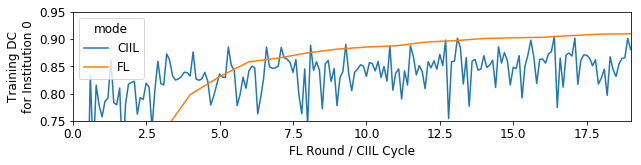}
\setlength{\abovecaptionskip}{-3pt}
\setlength{\belowcaptionskip}{-25pt}
\caption{CIIL catastrophic forgetting: first institution's training $DC$.}
\label{fig5}
\end{center}
\end{figure}

For our CIIL results in Fig.\ref{fig4}A, we show the distribution of scores after every cycle, rather than the best scores of a given run, to emphasize the importance of validation during training: CIIL does not support proper validation during training, as the model is not synchronized across the institutions until training is completed.

Despite the significant imbalance in numbers of subjects per institution (Table~\ref{table1}), FL achieves 98.7\% of the centralized validation $DC$, as do the best CIIL results. However, CIIL is less stable (Fig.\ref{fig4}), with a wide range of scores after each cycle. The instability of CIIL means that to learn a good model with CIIL, the model must be evaluated after each cycle. However, to evaluate the model, each institution must receive a copy of the model to test against its validation data. This adds additional communication overhead (i.e., an institution must receive the model 1 extra time per cycle) and requires a method to aggregate the results, at which point CIIL becomes arguably more complex than FL and with greater communication cost.

IIL learns a relatively poor model, averaging only 93\% of the validation $DC$, and suffers similar instability as CIIL. For our IIL experiments, each institution trained until there was no improvement in validation $DC$ for eight epochs (as measured by its own validation data), passing the best-performing model to the next institution.

Evaluation of the training data $DC$ scores for institution 0 during CIIL and FL training reveals that the CIIL models suffer from some amount of catastrophic forgetting \cite{catastrophicForgetting,overcomingForgetting}, i.e., the model "forgets" some of what it learned from the earlier institutions (Fig.\ref{fig5}). We verified that the peaks in the training $DC$ for CIIL indeed correspond to immediately after institution 0 trained the model. Forgetting could cause the instability we see in CIIL. We leave further investigation to future work. By comparison, FL maintains the training $DC$ for institution 0 throughout its training.

\subsection{Results for Random Simulated Distributions}
\begin{figure}
\begin{center}
\includegraphics[width=\textwidth]{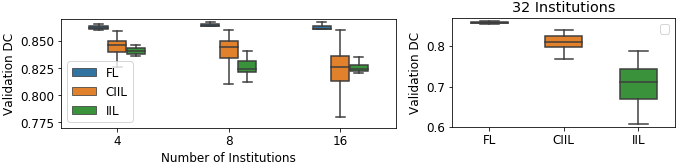}
\setlength{\abovecaptionskip}{-1pt}
\setlength{\belowcaptionskip}{-18pt}
\caption{Comparing FL, CIIL and IIL for collaborations of 4-32 institutions. Note that 32 simulations have a different y-axis range.}
\label{fig6}
\end{center}
\end{figure}

Figure~\ref{fig6} shows comparative results across FL, IIL and CIIL for the simulated data distributions, where, each institution was assigned roughly the same number of subjects (no subject's data was split across institutions), so they were far more balanced than the actual distribution. Note that in Fig.\ref{fig6}, the y-axis scale is different for 32 institutions, as the CIIL and IIL results were quite poor.

Our FL results show remarkable consistency on the simulated distributions, achieving 99+\% of the data-sharing results in all simulations, and FL also achieves superior results when compared to the best CIIL models. Even with 32 institutions, where each institution averaged fewer than 6 subjects, FL trains efficiently. In contrast, CIIL and IIL show considerable instability, with standard deviations ~10x that of FL for 16 and 32 institutions.

\begin{figure}
\begin{center}
\includegraphics[width=\textwidth]{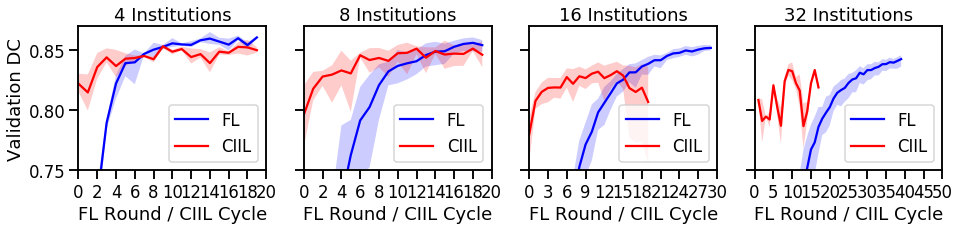}
\setlength{\abovecaptionskip}{-1pt}
\setlength{\belowcaptionskip}{-18pt}
\caption{FL and CIIL over round / cycle (1 epoch per). Confidence intervals are 0-100\%. Note that 16 and 32 institutions are shown to 30 and 50 rounds, respectively.}
\label{fig7}
\end{center}
\end{figure}

Figure~\ref{fig7} shows that for FL, while different numbers of institutions converge to similar model quality, they do not converge at the same rate. Note the different x-axis scales for 16 and 32 institutions. The causes for this are two-fold. First, with less data per institution, the model deltas are smaller at each round. Second, though the data is randomly distributed, the per-institution datasets become small enough that the individual institutions' datasets are less similar.

\begin{figure}
\begin{center}
\includegraphics[width=\textwidth]{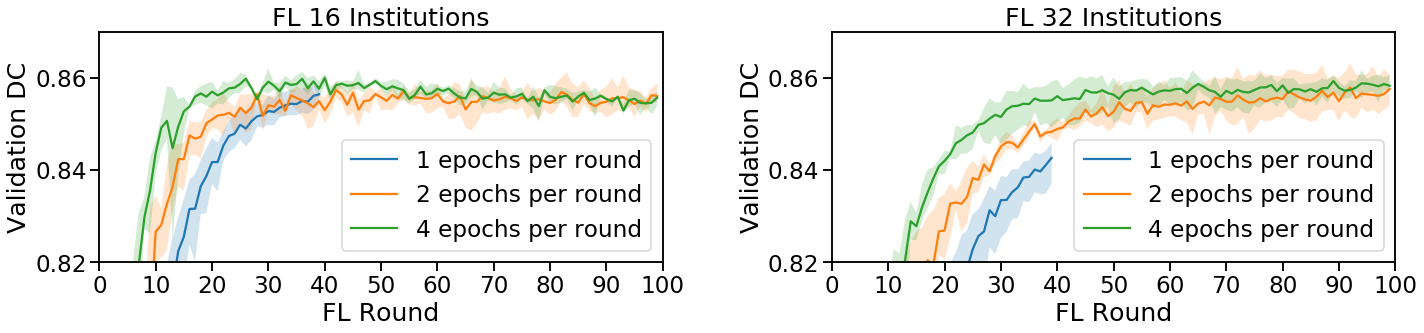}
\setlength{\abovecaptionskip}{-1pt}
\setlength{\belowcaptionskip}{-18pt}
\caption{FL over rounds for various EpR (16 and 32 institutions). Min/max shading.}
\label{fig8}
\end{center}
\end{figure}

By comparing FL experiments for 16 and 32 institutions at various EpR (Fig.\ref{fig8}), we note a convergence slowdown caused by smaller model deltas. The convergence slowdown is not quite proportional to the decrease in epochs, especially for 16 institutions.

\subsection{Hyper-Parameters}
All our experiments used a batch size of 64 and learning rate 5e-4 (Adam optimizer). We developed several solutions for adapting Adam for FL, all of which worked equivalently well in this domain. We leave investigating these options in a broader set of domains for future work.

\section{Practical considerations}

\subsection{Data Pre-processing}
Although the data is not centrally shared in FL, sources of variation across equipment configurations and acquisition protocols need to be considered. The uncontrolled varying acquisition environment of standard clinical practice, where the highest throughput of medical images is produced, make such data of limited use and significance in large-scale analytical studies, whereas data from more controlled environments (such as clinical trials) are more suitable. Since standardization of the acquisition protocols cannot be controlled, the pre-processing approaches should account for harmonization of heterogeneous data, allowing for integration and facilitating easier multi-institutional collaboration for large-scale analytics.

\subsection{Data Labeling Protocol}
The labeling protocol is instrumental to enable appropriate training of a ML model, allowing linking to reproducible expert clinical knowledge, while avoiding operator bias. Specifically, the definition and documentation of semantic descriptors of distinct anatomical regions is essential to allow reproducibility across institutions.

\subsection{Addition/Removal of Collaborators}
Institutions could be added, or removed, after some time of training, in any of the above collaborative learning configurations (FL, IIL, CIIL). In such cases, the model resulting from further collaborative training is expected to be qualitatively similar (after a transition period) to one obtained by training from scratch with the new set of collaborators. Eventually, any missing data would be forgotten. New data patterns will be learned subject to the limitations observed in this paper of the particular collaborative configuration, in face of the new data distribution. We leave such studies, however, to future work.

\section{Conclusions}
Our experiments demonstrate that the collaborating clinical institutions could train a model without sharing their data, using federated learning (FL). Our FL experiments achieve ~99\% of the model performance of a data-sharing model even with imbalanced datasets, such as the actual BraTS institutional distribution, or relatively few samples per participant, such as our simulation of 32 institutions with ~6 subjects per institution. While CIIL may seem a simpler alternative, in order to select a good model, full validation must be run often, such as at the end of each cycle. These validations would require the same synchronization and aggregation steps as FL, and would even add communication costs above FL. Finally, IIL and CIIL do not scale well to large number of institutions with small amounts of data.

Translation and adoption of such a FL system in a clinical configuration for multi-institutional collaboration, towards producing computer-aided analytics and assistive diagnostics, is expected to have a catalytic impact towards precision medicine, especially since introducing knowledge from another institution would improve the performance of the trained models without the need to share patient data, thereby overcoming potential privacy or data ownership concerns.

\section{Acknowledgements}
Research reported in this publication was partly supported by the National Institutes of Health (NIH) under award numbers NIH/NINDS:R01NS042645 and NIH/NCI:U24CA189523. The content of this publication is solely the responsibility of the authors and does not necessarily represent the official views of the NIH.

\bibliographystyle{splncs04}

\end{document}